\documentclass[sigconf]{acmart}

\renewcommand\footnotetextcopyrightpermission[1]{}
\setcopyright{none}

\acmConference[]{}{}{}
\acmBooktitle{}
\acmDOI{}
\acmISBN{}

\AtBeginDocument{
  
}
\usepackage{microtype}
\usepackage{tikz}
\usetikzlibrary{fit,positioning,backgrounds}
\usetikzlibrary{shapes.geometric}
\usepackage{hyperref}
\usepackage{amsmath}
\usepackage{mathtools}
\usepackage{amsthm}
\usepackage[textsize=tiny]{todonotes}

\usepackage[capitalize,noabbrev]{cleveref}

\theoremstyle{plain}

\theoremstyle{definition}

\theoremstyle{remark}

\usetikzlibrary{arrows.meta,positioning}
\usepackage{graphicx}
\usepackage{subcaption}
\usepackage{booktabs}
\usepackage{multirow}

\renewcommand\footnotetextcopyrightpermission[1]{}
\acmConference[]{}{}{}
\acmBooktitle{}
\acmDOI{}
\acmISBN{}
\title{\textsc{lever}: Inference-Time Policy Reuse under Support Constraints}

\author{Ihor Vitenko}
\affiliation{
  \institution{Anhalt University of Applied Sciences}
  \city{}
  \country{Germany}
}

\author{Noha Ibrahim}
\affiliation{
  \institution{Universit\'e Grenoble Alpes}
  \city{Grenoble}
  \country{France}
}

\author{Sihem Amer-Yahia}
\affiliation{
  \institution{CNRS, Univ. Grenoble Alpes}
  \city{Grenoble}
  \country{France}
}

\begin{document}
\fancyhead{}
\renewcommand{\headrulewidth}{0pt}
\begin{abstract}
Reinforcement learning (RL) policies are typically trained for fixed objectives,
making reuse difficult when task requirements change.
We study inference-time policy reuse: given a library of pre-trained policies
and a new composite objective, can a high-quality policy be constructed
entirely offline, without additional environment interaction?

We introduce \textsc{lever} (\textsc{L}everaging \textsc{E}fficient \textsc{V}ector
\textsc{E}mbeddings for \textsc{R}eusable policies), an end-to-end framework that
retrieves relevant policies, evaluates them using behavioral embeddings, and
composes new policies via offline Q-value composition.
We focus on the support-limited regime, where no value propagation is possible,
and show that the effectiveness of reuse depends critically on the coverage of
available transitions.

To balance performance and computational cost, \textsc{lever} proposes
composition strategies that control the exploration of candidate policies.
Experiments in deterministic GridWorld environments show that inference-time
composition can match, and in some cases exceed, training-from-scratch
performance while providing substantial speedups.
At the same time, performance degrades when long-horizon dependencies require
value propagation, highlighting a fundamental limitation of offline reuse.

Code is available\footnote{\url{https://github.com/Strongich/LEVER}}.
\end{abstract}
\maketitle

\section{Introduction}

Reinforcement learning (RL) policies are typically trained for fixed objectives.
When task requirements change, even slightly, a new policy often must be trained
from scratch, making RL systems expensive to deploy and limiting policy reuse.
Yet many real-world tasks are inherently compositional: an agent may need to
reach a goal, collect rewards, and avoid hazards, often in different combinations.
This raises a fundamental question: \emph{can pre-trained policies be reused to
solve new composite tasks without additional environment interaction?}

In this work, we study inference-time policy reuse, where a fixed library
of pre-trained policies must be leveraged to solve a new task entirely offline. It builds on the Model Reusability framework for RL~\cite{DBLP:journals/vldb/NikookarNRAO25}.
This setting is attractive for deployment, but also highly constrained: no new
data can be collected, and all decisions must rely on previously observed
behavior. In particular, reuse is limited by the \emph{support} of the policy
library, i.e., the set of state--action transitions that have been observed and
stored. Since no new interaction is allowed, missing transitions cannot be
recovered at inference time. The key challenge is therefore not only how to
compose policies, but when such reuse is possible at all.

A central insight of this paper is that inference-time reuse critically depends
on the underlying regime of the problem. When no value propagation is required
($\gamma = 0$), reuse is \emph{support-limited}: the quality of a composed policy
depends entirely on whether the necessary transitions are already present in the
library. In contrast, when long-horizon planning is required ($\gamma > 0$),
reuse becomes significantly more challenging, as optimal behavior depends on
propagating value across transitions that may not be jointly covered by the
available policies. This distinction fundamentally changes the nature of the
problem, shifting it from selection over known behaviors to implicit planning
over incomplete support.
This work builds on prior research on model reusability in RL~\cite{DBLP:journals/vldb/NikookarNRAO25}
and behavioral embeddings such as $\pi^2$VEC~\cite{pi2vec2024}, extending these ideas
to a practical inference-time setting where no additional interaction is allowed.
We introduce \textsc{lever}, an end-to-end
framework for inference-time policy reuse. Given a natural-language task
description and a library of pre-trained policies, \textsc{lever} retrieves
relevant candidates, evaluates them offline using behavioral embeddings, and
optionally composes new policies without interacting with the environment.
Importantly, \textsc{lever} serves as a practical instrument to explore the
limits of reuse under different regimes, rather than relying on any specific
composition operator.

A key practical challenge is how broadly to explore candidate policies under
these constraints. Composing too few policies may miss critical transitions,
while exploring too many combinations quickly becomes computationally
intractable. To study this trade-off, we consider three composition strategies
with different performance scalability trade-offs: \emph{Targeted Composition},
which selects a single policy per subtask; \emph{Exhaustive Composition}, which
enumerates all combinations; and \emph{Hybrid Composition}, which explores a
restricted set of top candidates. These strategies allow us to analyze how
selection and combinatorial search interact with the underlying reuse
constraints.

We evaluate inference-time reuse across both support-limited and
planning-enabled settings, in tabular and deep RL environments.
Our results show that reuse is highly effective in the support-limited regime,
where simple composition strategies can match or exceed training-from-scratch
performance while providing substantial computational savings.
However, when applied outside this regime, performance degrades and becomes
sensitive to coverage and evaluation errors, highlighting the limitations of
purely offline reuse.

Overall, our findings suggest that the main bottleneck of inference-time policy
reuse is not the composition mechanism itself, but the availability of sufficient
support to enable reliable decision-making. This provides a clearer understanding
of when reuse can serve as a practical alternative to retraining, and where its
limitations lie.

\section{Related Work}
\label{sec:relwork}

We study inference-time policy reuse in reinforcement learning: given a fixed
library of pre-trained policies, can new tasks be solved entirely offline,
without additional environment interaction?
Our work is most closely related to research on policy reuse and model
reusability in RL, but differs in its strict offline setting and its focus on
characterizing when reuse is feasible.

\noindent \emph{Policy Reuse and Transfer in RL.}
A large body of work investigates how knowledge can be transferred across tasks.
Multi-task RL learns shared representations across objectives~\cite{vithayathil2020survey},
while meta-RL enables rapid adaptation to new tasks by training over task
distributions~\cite{gupta2018meta}.
Other approaches reuse value functions, initialize policies, or rely on modular
architectures and curricula~\cite{singh1992transfer,taylor2009transfer,
singh1996reinforcement,sutton2018reinforcement,barrett2010transfer}.
These methods are effective, but crucially depend on additional training or
environment interaction when objectives change.

In contrast, we consider a strictly inference-time setting: the policy library
is fixed, and no new data can be collected.
This setting removes the possibility of adaptation and shifts the problem from
learning to \emph{selection and composition under limited support}.
Our work builds on the Model Reusability framework for RL~\cite{DBLP:journals/vldb/NikookarNRAO25},
which provides theoretical conditions under which policies can be reused without
retraining.
We extend this perspective by empirically studying how reuse behaves under
different regimes and by introducing practical mechanisms for retrieval,
evaluation, and composition.

\noindent \emph{Model Reusability and Offline Decision-Making.}
More broadly, our work relates to settings where decisions must be made without
additional data collection.
In offline RL, policies are learned from fixed datasets without further
interaction, raising challenges related to coverage and distributional
shift~\cite{sutton2018reinforcement}.
Similarly, case-based reasoning and experience reuse methods leverage past
trajectories to guide decision-making~\cite{von2005abstracting,yu2018towards,yu2018reusable}.
However, these approaches typically focus on learning from data, whereas we
assume access to a library of \emph{policies} and study how they can be reused
directly at inference time.

Our work is also related to broader notions of model reuse in machine learning,
such as pre-trained language models~\cite{devlin2018bert}, federated learning~\cite{mcmahan2017communication},
and AutoML systems~\cite{vanschoren2014openml}, where previously trained models
are adapted or combined for new tasks.
In contrast, we focus on the reuse of decision policies in sequential settings,
and highlight how the absence of interaction fundamentally constrains what can
be achieved.

\section{Problem Setting and Solution Overview}

We consider a reinforcement learning setting modeled as a Markov Decision
Process (MDP) with state space $\mathcal{S}$, action space $\mathcal{A}$,
transition dynamics $\mathcal{P}$, reward function $\mathcal{R}$, and discount
factor $\gamma \in [0,1)$.
We assume a model-free setting~\cite{sutton2018reinforcement}, where transition
dynamics are not accessible at inference time and no additional interaction with
the environment is allowed.

\paragraph{Inference-time reuse setting.}
We are given a fixed library of pre-trained policies
$\{\pi_1, \ldots, \pi_n\}$, each optimized for a base reward function
$\{R_1, \ldots, R_n\}$ defined over a shared MDP.
All policies operate over the same state and action spaces and can be queried
through a common interface $a \sim \pi_i(\cdot \mid s)$.
At inference time, the policy library is fixed: no new data can be collected,
and no additional training is performed.

\paragraph{Composite objectives.}
We consider new tasks whose reward function is expressed as a linear combination
of base rewards:
\begin{equation}
R_{\text{new}}(s,a,s') = \sum_{i=1}^{n} w_i R_i(s,a,s'),
\end{equation}
where $w_i \in \mathbb{R}$ are importance weights.
In practice, the number of relevant policies is determined dynamically through
task decomposition and retrieval, and we adopt a uniform weighting
($w_i = 1$) in our experiments to isolate the effect of reuse and composition.

Given a composite objective $R_{\text{new}}$ and a policy library, our goal is
to construct a policy $\pi_{\text{new}}$ that performs competitively with a
policy trained from scratch on $R_{\text{new}}$, while requiring no
additional interaction with the environment.

\paragraph{Two regimes of reuse.}
A central aspect of this problem is the role of the discount factor $\gamma$,
which determines whether value propagation is required.

When $\gamma = 0$, the return depends only on immediate rewards, and policy
evaluation is \emph{support-limited}: the value of a policy depends entirely on
the transitions already observed in the library.
In this regime, reuse reduces to selecting or combining behaviors over known
transitions, without requiring long-horizon reasoning.

In contrast, when $\gamma > 0$, optimal behavior depends on value propagation
across multiple steps.
This introduces a planning component: composing policies requires combining
transition structures and propagating value estimates across states that may not
be jointly covered by any single policy.
As a result, reuse becomes sensitive to coverage, consistency, and approximation
errors.

This distinction between support-limited ($\gamma = 0$) and planning-enabled
($\gamma > 0$) regimes fundamentally changes the nature of inference-time reuse
and serves as the backbone of our analysis.

This work builds on prior research on model reusability in RL~\cite{DBLP:journals/vldb/NikookarNRAO25}
and behavioral embeddings such as $\pi^2$VEC~\cite{pi2vec2024}, extending these ideas
to a practical inference-time setting where no additional interaction is allowed.

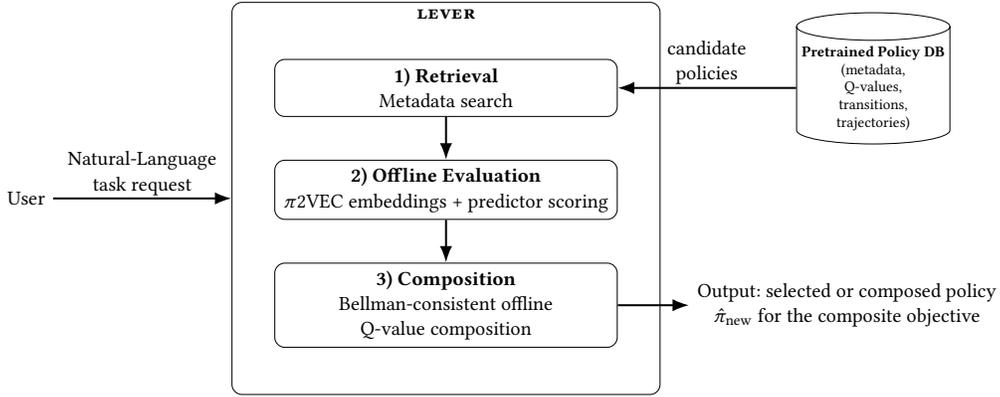
\begin{figure*}[h]
\centering
\resizebox{0.75\linewidth}{!}{%
\begin{tikzpicture}[
    font=\small,
    node distance=8mm and 10mm,
    bigbox/.style={draw, rounded corners, align=center, minimum width=60mm, minimum height=55mm, inner sep=6pt},
    layer/.style={draw, rounded corners, align=center, minimum width=48mm, minimum height=8mm, inner sep=3pt},
    db/.style={
        draw,
        cylinder,
        shape border rotate=90,
        aspect=0.18,
        align=center,
        minimum height=12mm,
        minimum width=16mm,
        inner sep=2pt,
        font=\scriptsize
    },
    arrow/.style={-Latex, thick}
]

\node[draw=none, fill=none] (user) {User};

\node[bigbox, right=25mm of user] (lever) {};

\node[anchor=north] at (lever.north) {\textbf{\textsc{lever}}};

\node[layer, anchor=north] (retr) at ([yshift=-8mm]lever.north)
{\textbf{1) Retrieval}\\Metadata search};

\node[layer, below=6mm of retr] (embed)
{\textbf{2) Offline Evaluation}\\\(\pi\)2VEC embeddings + predictor scoring};

\node[layer, below=6mm of embed] (comp)
{\textbf{3) Composition}\\Bellman-consistent offline\\ Q-value composition};

\node[db, right=25mm of retr] (DB)
{\textbf{Pretrained Policy DB}\\
(metadata,\\ Q-values,\\ transitions,\\ trajectories)};

\draw[arrow] (user.east) -- node[above, align=center, inner sep=1pt, font=\small]
{Natural-Language\\ task request} (lever.west);

\draw[arrow] (DB.west) -- node[above, align=center, inner sep=1pt]
{candidate \\ policies} (retr.east);

\draw[arrow] (retr.south) -- (embed.north);
\draw[arrow] (embed.south) -- (comp.north);

\draw[arrow] (comp.east) -- ++(10mm,0mm)
node[right, align=center]
{Output: selected or composed policy\\\(\hat{\pi}_{\text{new}}\) for the composite objective};

\end{tikzpicture}%
}
\caption{High-level overview of \textsc{lever}. A user specifies a task in natural language. \textsc{lever} retrieves relevant policies from a pretrained policy database, evaluates candidates offline using \(\pi\)2VEC embeddings, and composes policies offline when needed.}
\label{fig:lever_overview}
\end{figure*}

\paragraph{Solution overview.}
To study inference-time reuse in practice, we introduce \textsc{lever}, a modular
framework for retrieving, evaluating, and composing policies entirely offline
(Fig.~\ref{fig:lever_overview}).

\textsc{lever} takes as input (i) a natural-language task description and
(ii) a policy library augmented with metadata and offline artifacts (e.g.,
trajectories, Q-values, transition graphs, and reward functions).
It outputs either a selected policy from the library or a composed policy
$\hat{\pi}_{\text{new}}$ for the new task.

The framework operates in three stages:
(1) \emph{Retrieval}: the task description is decomposed into subtasks and used
to retrieve relevant policies via semantic matching over metadata;
(2) \emph{Offline evaluation}: candidate policies are represented using
behavioral embeddings derived from successor features ($\pi^2$VEC), and ranked
by a predictor estimating their expected performance;
(3) \emph{Composition}: when no single policy is sufficient, new policies are
constructed by combining Q-values and transition graphs under the composite
reward.

All steps are performed offline using stored artifacts, without additional
rollouts or interaction with the environment.

\paragraph{Composition strategies.}
Within this framework, we consider three strategies that differ in how broadly
composition is explored.
\emph{Targeted Composition (TC)} selects the single most promising policy per
subtask and composes only these candidates.
\emph{Exhaustive Composition (EC)} enumerates a large set of policy combinations,
incurring significant computational cost.
\emph{Hybrid Composition (HC)} interpolates between these extremes by selecting
a small set of top-$k$ candidates per subtask before composing them, providing
a controllable trade-off between performance and scalability.

These strategies allow us to analyze how the effectiveness of reuse depends on
the underlying regime and the extent of combinatorial exploration.
Sections~\ref{sec:composition} and~\ref{sec:tc-ec} provide detailed formulations.

\section{\textsc{lever} Framework}

We present \textsc{lever}, a modular framework for inference-time policy reuse.
Its goal is to operationalize the offline reuse setting introduced in the
previous section, by providing a concrete pipeline for retrieving, evaluating,
and composing policies without any environment interaction.

Given a natural-language task description and a library of pre-trained policies,
\textsc{lever} proceeds in three stages.
First, it retrieves a set of candidate policies relevant to the task by matching
the task description with policy metadata.
Second, it evaluates these candidates offline using behavioral embeddings based
on successor features ($\pi^2$VEC), enabling comparison without execution.
Finally, when no single policy is sufficient, it constructs new policies by
combining existing ones through offline composition operators.

Importantly, all components of \textsc{lever} operate under the same constraint:
no additional data is collected, and all decisions rely solely on previously
observed policy artifacts.
As such, the framework serves as a practical tool to study how different reuse
strategies behave under the support-limited and planning-enabled regimes
introduced earlier.

In the following sections, we detail the offline evaluation mechanism and the
policy composition strategies used within \textsc{lever}.

\subsection{Policy Retrieval}
\label{sec:retrieval}

Policy retrieval selects a small set of candidate policies from the library that
are relevant to a given task.
In the inference-time setting, this step is critical: since no additional
interaction is allowed, all subsequent evaluation and composition are restricted
to the retrieved policies.
Retrieval therefore directly determines the available support for reuse.

The input task is specified in natural language and may involve multiple
objectives.
We decompose this instruction into a set of subtasks $\{\tau_k\}$, each capturing
a distinct objective.
For example, the instruction \textit{``Find the fastest exit and collect as much
gold as possible''} can be decomposed into
$\tau_1=\text{reach the exit quickly}$ and
$\tau_2=\text{maximize gold collection}$.
Unless specified otherwise, all subtasks are treated with equal importance.

Each pre-trained policy is associated with metadata describing its training
objective and environment.
These metadata entries are embedded into a shared vector space and indexed in a
vector database (VDB).\footnotemark\ 
For each subtask $\tau_k$, its textual description is embedded and used to query
the VDB, returning a ranked list of candidate policies $C_k$ based on similarity.
Retrieval is performed independently for each subtask.

The retrieved candidates are merged into a base policy pool
\begin{equation}
\mathcal{P}_{\text{base}} = \bigcup_k C_k,
\end{equation}
which defines the set of policies available for offline evaluation and
composition.
In practice, $\mathcal{P}_{\text{base}}$ is small (on the order of a few to tens
of policies), enabling tractable downstream processing.

Importantly, this pool also defines the \emph{support} of the reuse process:
if relevant behaviors are not present in $\mathcal{P}_{\text{base}}$, no
subsequent composition step can recover them.
As a result, retrieval plays a central role in determining when inference-time
reuse succeeds or fails.
\footnotetext{In our implementation, we use a lightweight LLM for task decomposition and embedding.}

\subsection{Offline Q-Value Composition}
\label{sec:composition}

Policy composition at inference time depends critically on whether long-horizon
value propagation is required.
Following~\cite{DBLP:journals/vldb/NikookarNRAO25}, we distinguish two regimes.

\paragraph{Two regimes of composition.}
In the \emph{planning-enabled regime} ($\gamma > 0$), optimal composition
requires Bellman propagation over the union of transition graphs.
In principle, this allows recovery of optimal policies, but in practice it
requires reasoning over large, possibly cyclic graphs and relies on strong
coverage assumptions.
As a result, fully offline composition in this regime is computationally
expensive and sensitive to missing transitions.

In contrast, in the \emph{support-limited regime} ($\gamma = 0$), no value
propagation is performed.
Q-values depend only on immediate rewards, and composition reduces to combining
existing estimates over observed transitions.
In this case, the quality of the composed policy is entirely determined by the
availability of relevant transitions in the policy library.

In this work, we focus on the support-limited regime as the primary setting for
scalable inference-time reuse.
The planning-enabled formulation is included for comparison.

\paragraph{Setting.}
We assume a set of base policies trained on a shared MDP.
For each policy $\pi_i$, we have access to an action-value function $Q_i(s,a)$
and a set of observed transitions
$\mathcal{T}_i \subseteq \mathcal{S} \times \mathcal{A} \times \mathcal{S}$.
We consider deterministic environments, where each $(s,a)$ leads to a unique
successor state.

\paragraph{Composite reward.}
Given weights $w \in \mathbb{R}^n$, the composite reward is defined as
\begin{equation}
R_{\text{new}}(s,a,s') = \sum_{i=1}^n w_i\, R_i(s,a,s').
\label{eq:rnew}
\end{equation}

\paragraph{Shared support.}
Since composition is performed without interaction, it is restricted to
transitions that are available in all selected policies.
Let $\mathcal{U}_i = \{(s,a) \mid (s,a,s') \in \mathcal{T}_i\}$ denote the
support of policy $\pi_i$.
Given a subset of policies selected for composition (Section~\ref{sec:tc-ec}),
we define the shared support as:
\begin{equation}
\mathcal{U}_{\cap} = \bigcap_{i=1}^n \mathcal{U}_i.
\label{eq:ucap}
\end{equation}

Composition is restricted to $(s,a) \in \mathcal{U}_{\cap}$, ensuring that
Q-values are combined only over transitions that are jointly observed.
This constraint reflects the core limitation of inference-time reuse: if a
transition is missing from the selected policies, it cannot be recovered by
composition.

\paragraph{Support-limited composition.}
In the support-limited regime ($\gamma = 0$), Q-values depend only on immediate
rewards, and composition reduces to combining base Q-values over the shared
support.
This avoids Bellman propagation and enables efficient offline computation,
but makes performance entirely dependent on coverage.

\paragraph{Out-of-regime evaluation.}
In addition to the theoretically grounded setting, we also evaluate composition
using policies trained with $\gamma = 0.99$ while still applying the
support-limited operator at inference time.
This setting falls outside the assumptions of
\cite{DBLP:journals/vldb/NikookarNRAO25} and does not provide optimality
guarantees.
Instead, it serves as an empirical probe of how composition behaves when
long-horizon effects are present but cannot be propagated.

\subsubsection{Composing Q-values}
\label{subsec:compose-q}

\paragraph{Support-limited regime ($\gamma = 0$).}
The composed Q-function is initialized as
\begin{equation}
Q_{\text{new}}(s,a)=\sum_{i=1}^n w_i\, Q_i(s,a),
\qquad (s,a)\in\mathcal{U}_{\cap}.
\label{eq:qinit}
\end{equation}
In this regime, no Bellman backup is performed, and Q-values depend only on
immediate rewards.
This corresponds to the \textsc{ExZeroDiscount} formulation
from~\cite{DBLP:journals/vldb/NikookarNRAO25}.
As a result, composition is efficient and fully offline, but strictly limited
by transition coverage.

\paragraph{Planning-enabled regime ($\gamma > 0$).}
When $\gamma > 0$, the optimal composed Q-function satisfies
\begin{equation}
Q_{\text{new}}(s,a)=\sum_{i=1}^n w_i R_i(s,a)
+\gamma\max_{a'}Q_{\text{new}}(s',a').
\label{eq:planning_regime}
\end{equation}
Solving this equation requires Bellman propagation over the union of transition
graphs.
While this can in principle recover optimal behavior, it involves combinatorial
search and relies on strong coverage assumptions, making it impractical for
fully offline reuse at scale.
For this reason, our main experiments focus on the support-limited regime.

\paragraph{Policy extraction.}
Given a composed Q-function, the resulting policy is extracted greedily:
\begin{equation}
\pi_{\text{new}}(s)=\arg\max_a Q_{\text{new}}(s,a).
\label{eq:pi-new}
\end{equation}

\paragraph{Deep RL composition via GPI}
For deep RL policies, direct Q-value summation is not always applicable, as
policies may not share a common tabular representation.
Instead, we rely on Generalised Policy Improvement (GPI)~\cite{barreto2019transferdeepreinforcementlearning}
as an approximate composition mechanism.

For DQN, each policy exposes per-action Q-values, and the composed action is:
\begin{equation}
\pi_{\text{new}}(s) = \arg\max_{a} \max_{i} Q_i(s, a).
\label{eq:gpi_dqn}
\end{equation}

For PPO, policies do not expose per-action Q-values.
Instead, each policy provides a state-value estimate $V_i(s)$, and action
selection is delegated to the policy with the highest value:
\begin{equation}
\pi_{\text{new}}(s) = \pi_{i^*}(s), \qquad i^* = \arg\max_{i} V_i(s).
\label{eq:gpi_ppo}
\end{equation}

These compositions are performed entirely offline using stored model outputs,
but unlike the tabular setting, they do not provide exact Q-value composition
and should be viewed as approximations.

\paragraph{Candidate selection.}
Composition is applied to subsets of policies selected by the different
strategies (Targeted, Exhaustive, or Hybrid).
The resulting candidate policies are then evaluated offline using the
embedding-based predictor (Section~\ref{sec:policy-evaluation}) to select the
final policy $\widehat{\pi}$.

\subsection{Composition Strategies}
\label{sec:tc-ec}

After policy retrieval, the framework must select or construct a policy that
optimizes the composite objective using only the retrieved candidates.
Since all decisions are made offline, the search for a suitable policy is
restricted to combinations of the available policies, and the main design choice
is how broadly this space is explored.

We consider three composition strategies that differ in the extent to which they
explore the available support.
These strategies do not change the underlying composition operator, but control
how many candidate policies are considered and combined.
As a result, they define a trade-off between computational cost and the ability
to recover useful behaviors from the policy library.

At one extreme, strategies that consider very few candidates are efficient but
may miss important combinations.
At the other extreme, exhaustive exploration can recover stronger solutions but
quickly becomes computationally intractable.
Between these two, intermediate strategies aim to balance coverage and cost.

We describe these strategies in increasing order of search breadth, and use them
to analyze how the effectiveness of inference-time reuse depends on the extent
of combinatorial exploration.

\subsubsection{Targeted Composition (TC)}
\label{subsec:tc}

Targeted Composition (TC) represents the most restrictive form of exploration.
It selects a single candidate policy for each subtask and composes only this
minimal set.

Concretely, all retrieved policies are evaluated offline using the predictor
$f(\Psi_\pi)$.
For each subtask, only the top-ranked policy is retained, and the selected
policies are composed using the operator described in
Section~\ref{sec:composition}.
The resulting composed policy $\widehat{\pi}$ is returned.

By limiting composition to one policy per subtask, TC minimizes computational
cost and enables fast inference-time reuse.
However, it relies entirely on the quality of the initial ranking: if a relevant
policy is not selected, it cannot be recovered through composition.
As a result, TC is highly sensitive to retrieval and evaluation errors, and may
fail when useful behaviors are not captured by the top-ranked candidates.

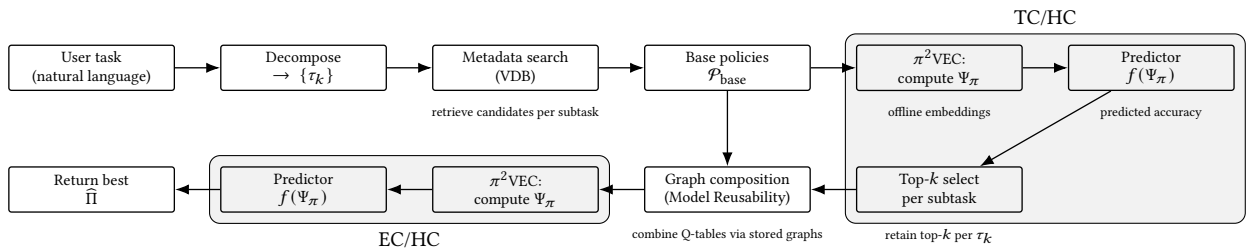
\begin{figure*}[h]
\centering
\begin{tikzpicture}[
  font=\scriptsize,
  box/.style={
    draw, rounded corners=1pt,
    minimum height=6mm,
    minimum width=22mm,
    align=center,
    inner sep=1.5pt,
    line width=0.5pt
  },
  note/.style={font=\tiny, align=center},
  arr/.style={-Latex, line width=0.5pt}
]

\node[box] (user) {User task\\(natural language)};
\node[box, right=6mm of user] (decomp) {Decompose\\$\rightarrow\ \{\tau_k\}$};
\node[box, right=6mm of decomp] (vdb) {Metadata search\\(VDB)};
\node[box, right=6mm of vdb] (base) {Base policies\\$\mathcal{P}_{\text{base}}$};
\node[box, right=6mm of base] (pi2)
  {$\pi^2$VEC:\\compute $\Psi_\pi$};
  \node[box, right=6mm of pi2] (pred)
  {Predictor\\$f(\Psi_\pi)$};
  
\draw[arr] (user) -- (decomp);
\draw[arr] (decomp) -- (vdb);
\draw[arr] (vdb) -- (base);
\draw[arr] (base) -- (pi2);

\node[note, below=1.0mm of vdb] {retrieve candidates per subtask};


\node[box, below=10mm of pi2] (topk)
  {Top-$k$ select\\per subtask};

  \begin{pgfonlayer}{background}
\node[draw,
      rounded corners,
      fill=gray!10,
      fit=(pi2)(pred)(topk),
      inner sep=4pt,
      label={[font=\small]above:TC/HC}] {};
\end{pgfonlayer}

\node[box, left=6mm of topk] (graph)
  {Graph composition\\(Model Reusability)};
\node[box, left=6mm of graph] (pi2_2)
  {$\pi^2$VEC:\\compute $\Psi_\pi$};
\node[box, left=6mm of pi2_2] (pred_2)
  {Predictor\\$f(\Psi_\pi)$};
\node[box, left=6mm of pred_2] (ret)
  {Return best\\$\widehat{\Pi}$};
\begin{pgfonlayer}{background}
\node[draw,
      rounded corners,
      fill=gray!10,
      fit=(pred_2)(pi2_2),
      inner sep=4pt,
      label={[font=\small]below:EC/HC}] {};
\end{pgfonlayer}

\draw[arr] (base) -- (graph);
\draw[arr] (pi2) -- (pred);
\draw[arr] (pred) -- (topk);
\draw[arr] (topk) -- (graph);
\draw[arr] (graph) -- (pi2_2);
\draw[arr] (pi2_2) -- (pred_2);
\draw[arr] (pred_2) -- (ret);

\node[note, below=1.0mm of pi2] {offline embeddings};
\node[note, below=1.0mm of pred] {predicted accuracy};
\node[note, below=1.0mm of topk] {retain top-$k$ per $\tau_k$};
\node[note, below=1.0mm of graph] {combine Q-tables via stored graphs};

\end{tikzpicture}%

\caption{\textsc{lever} execution pipeline. TC and HC restrict composition to selected base policies, while EC enumerates all combinations. HC and EC evaluate composed policies offline and select the best candidate.}
\label{fig:lever}
\end{figure*}

\subsubsection{Hybrid Composition (HC)}
\label{subsec:hc}

Hybrid Composition (HC) explores a limited but diverse subset of the available
support.
Instead of selecting a single policy per subtask, it retains multiple promising
candidates and composes them.

All retrieved policies are first evaluated offline using the predictor
$f(\Psi_\pi)$.
For each subtask, the top-$k$ policies are retained, forming a small candidate
set per objective.
These candidates are then composed across subtasks using the operator described
in Section~\ref{sec:composition}, producing a set of composed policies that are
evaluated offline.
The highest-scoring composed policy is selected.

By considering multiple candidates per subtask, HC increases the chance of
recovering useful combinations that are missed by Targeted Composition.
At the same time, restricting the search to the top-$k$ candidates avoids the
combinatorial cost of exhaustive exploration.
The parameter $k$ therefore directly controls the trade-off between coverage of
the support and computational cost.

\subsubsection{Exhaustive Composition (EC)}
\label{subsec:ec}

Exhaustive Composition (EC) explores the largest possible portion of the
available support by considering all combinations of candidate policies across
subtasks.

Given the retrieved policies, EC enumerates all possible selections, composes
each combination using the operator described in
Section~\ref{sec:composition}, and evaluates the resulting policies offline.
The highest-scoring composed policy is selected.

By exploring the full combinatorial space, EC can recover solutions that rely on
specific interactions between policies.
However, its computational cost grows exponentially with the number of subtasks
and candidates, making it impractical for large-scale settings.
EC therefore serves primarily as an upper bound on what can be achieved through
offline composition given the available support.

\subsection{Policy Evaluation via Behavioral Embeddings}
\label{sec:policy-evaluation}

A central challenge in inference-time policy reuse is to compare candidate
policies without interacting with the environment.
Since no additional data can be collected, evaluation must rely entirely on
offline information and approximate representations of policy behavior.

\paragraph{Policy embeddings.}
We represent each policy $\pi$ using a fixed-dimensional embedding
$\Psi_\pi$ derived from successor features via the $\pi^2$VEC framework.
These embeddings summarize how a policy behaves across a set of states by
encoding expected accumulations of structured state features $\phi(s)$.
Aggregating these representations over a fixed set of reference states yields a
compact behavioral description of the policy.

\paragraph{Offline performance prediction.}
To rank candidate policies, we use a learned predictor $f(\Psi_\pi)$ that
estimates their expected return from embeddings.
The predictor is trained offline using a dataset of policies, where each
embedding is paired with an empirical return computed from stored trajectories.
We use a histogram-based gradient boosting regressor, which provides sufficient
expressiveness while remaining efficient; training details are provided in
Section~\ref{subsec:predictor-training}.

This predictor does not provide exact performance estimates, but rather an
approximate ranking signal based on behavioral similarity and past outcomes.

\paragraph{Offline data and selection.}
All embeddings and predictions rely exclusively on offline data.
For stored policies, we use trajectories collected during training; for composed
policies, trajectories are generated from their Q-functions when available.

Given a candidate set $\mathcal{C}$, we select the policy with the highest
predicted score:
\[
\widehat{\pi}
=
\arg\max_{\pi \in \mathcal{C}} f(\Psi_\pi).
\]

This procedure enables scalable comparison of both base and composed policies
without environment interaction, but its effectiveness depends on how well the
offline data captures the relevant behaviors.

\section{Experiments}
\label{sec:experiments}

Our experiments aim to understand when inference-time policy reuse is effective
under the constraints defined in Section~3.
In particular, we study how performance depends on the underlying regime
($\gamma = 0$ vs.\ $\gamma > 0$) and on the extent to which the available
policy space is explored.

We evaluate \textsc{lever} along two axes:
(i) \emph{policy quality} under a composite objective, and
(ii) \emph{offline computation cost} required to produce a policy.
Policy quality is measured relative to a baseline that trains a policy from
scratch on the composite reward, referred to as
\emph{Training-from-Scratch} (TFS).

We compare three composition strategies that explore the policy space at
different levels: Targeted Composition (TC), Hybrid Composition (HC), and
Exhaustive Composition (EC).
These strategies allow us to analyze how increasing the breadth of exploration
affects performance under the offline constraint.

Our evaluation is structured around two complementary settings.
First, we consider the \emph{support-limited regime} ($\gamma = 0$), where reuse
relies solely on the availability of relevant transitions in the policy
library.
Second, we study an \emph{out-of-regime setting} ($\gamma > 0$), where
long-horizon effects are present but cannot be propagated by the composition
operator.
This setting serves to assess how reuse behaves when its underlying assumptions
are violated.

Through this analysis, we aim to identify the conditions under which
inference-time reuse can match or approximate training from scratch, and to
highlight the limitations that arise when coverage or evaluation is
insufficient.

\subsection{Environment}
\label{subsec:env}

We evaluate our approach on deterministic GridWorld environments of size
$8\times8$ and $16\times16$.
Each environment is defined by a layout containing a start cell, an exit cell,
gold cells, obstacle (block) cells, hazard cells, and a lever cell.

A \emph{seed} corresponds to a randomly generated layout.
For each seed, we sample a single layout and reuse it across all reward
settings to ensure comparability.
To guarantee feasibility, we enforce that at least one hazard-free path exists
from start to exit and from start to lever to exit.
This is verified using breadth-first search (BFS); layouts that do not satisfy
these constraints are resampled.

Episodes terminate when the agent reaches the exit, steps on a hazard (terminal
failure), or after $N^2$ steps.

\paragraph{Reward settings.}
We define four base objectives: \texttt{path}, \texttt{gold},
\texttt{hazard}, and \texttt{lever}, each with dense shaping.
Composite tasks are constructed by summing subsets of these objectives
(e.g., \texttt{path-gold}, \texttt{path-gold-hazard},
\texttt{path-gold-hazard-lever}).

\paragraph{Action space.}
The agent has four deterministic actions (up, down, left, right).
Invalid moves (e.g., hitting a boundary or obstacle) leave the agent in place.

\subsection{Baseline and Experimental Settings}
\label{subsec:methods-compared}

We compare the three composition strategies (TC, HC, EC) to a baseline that
trains a policy from scratch on the composite reward, referred to as
\emph{Training-from-Scratch} (TFS).

\paragraph{Training-from-Scratch (TFS).}
In the tabular setting, TFS uses SARSA with learning rate $\alpha=0.1$.
Exploration follows an $\varepsilon$-greedy schedule decaying from $1.0$ to
$0.01$.
We consider three training budgets:
$X1$ (10k episodes), $X5$ (50k episodes), and $X10$ (100k episodes), with decay
rates adjusted to maintain comparable exploration profiles.

\paragraph{Experimental regimes.}
We evaluate reuse under two main settings.

\textbf{Support-limited regime ($\gamma = 0$).}
This is the primary setting of the paper.
Both the policy library and the TFS baseline are trained with $\gamma = 0$,
ensuring that value propagation is not required.
This corresponds to the regime where the composition operator is theoretically
well-founded~\cite{DBLP:journals/vldb/NikookarNRAO25}.
Experiments are conducted on $8\times8$ and $16\times16$ grids with 5 random
seeds per configuration.

\textbf{Out-of-regime setting ($\gamma > 0$).}
To assess how reuse behaves beyond its theoretical assumptions, we also consider
policies trained with $\gamma = 0.99$ while applying the same support-limited
composition operator at inference time.
This setting introduces long-horizon effects that cannot be propagated during
composition, and therefore serves as a stress test rather than a setting with
guarantees.

\paragraph{Deep RL settings.}
To evaluate the generality of the approach, we also consider deep RL policies.

\textbf{DQN.}
Policies are trained using DQN with a convolutional architecture on the
\texttt{LeverGrid} MiniGrid environment, using a local $7\times7$ observation.
Composition is performed via GPI (Eq.~\ref{eq:gpi_dqn}).
The TFS baseline corresponds to a monolithic DQN trained on the composite
reward.
Experiments are conducted on $8\times8$ and $16\times16$ grids.

\textbf{PPO.}
Policies are trained using PPO with the same architecture and observation
setup as DQN.
Composition is performed via critic-based GPI (Eq.~\ref{eq:gpi_ppo}).
The TFS baseline is a monolithic PPO policy trained on the composite reward.
Experiments are reported on the $8\times8$ grid.

Across all settings, composition is performed entirely offline using stored
policy artifacts, without additional environment interaction.

\subsection{Evaluation Metrics}
\label{subsec:metrics}

\paragraph{Policy quality.}
We measure policy quality using the \emph{average episodic return} of the final
policy over a set of evaluation instances.
This metric is used to compare inference-time reuse against the
Training-from-Scratch (TFS) baseline under the same composite objective.

\paragraph{Offline computation cost.}
We report the \emph{end-to-end wall-clock time} required to produce the final
policy.
For TFS, this corresponds to the full training time.
For TC, HC, and EC, this includes task decomposition, offline evaluation
(scoring), and policy composition.
This metric captures the computational advantage of inference-time reuse under
the no-interaction constraint.

\subsection{Policy Dataset and Diversity}
\label{subsec:policy-dataset}

To support offline evaluation and robust policy selection, we construct a
diverse dataset of policies that captures a wide range of behaviors.

Diversity is introduced along three axes:
(i) environment layouts (random seeds),
(ii) training progress, and
(iii) training budget.
Different seeds generate distinct deterministic GridWorld layouts
(e.g., variations in gold, hazards, obstacles, and lever placement),
leading to policies with different behaviors even under identical
hyperparameters.

For each training run, we select three reward-stratified checkpoints
(best, mid, low), capturing strong, average, and weak policies without relying
on fixed training iterations.
We further vary the exploration budget using three episode regimes
($X1$, $X5$, $X10$).

Across seeds, checkpoints, and budgets, this results in a diverse set of
policies for each reward type.
This diversity is essential for training the offline predictor and for enabling
meaningful comparisons between candidate policies under the limited-support setting.

\subsection{Offline Predictor Training}
\label{subsec:predictor-training}

To compare candidate policies without environment interaction, we train an
offline performance predictor
$f(\Psi_{\pi}) \rightarrow \widehat{J}(\pi)$ that maps a policy embedding
$\Psi_\pi$ to an estimate of its expected return.
This predictor is used to rank both retrieved base policies and composed
candidates.

\paragraph{Role of the predictor.}
Since no additional data can be collected at inference time, policy selection
must rely on approximate signals derived from offline information.
The predictor therefore provides a \emph{ranking signal} rather than an exact
performance estimate, allowing the system to compare candidates based on their
behavioral embeddings.

\paragraph{Avoiding temporal leakage.}
A key challenge is preventing \emph{future leakage}: the predictor must not be
trained on policies that would be unavailable at inference time.
To ensure a realistic evaluation setting, we train stage-specific predictors
using only the policies present in the library at each composition stage and
training budget.
This prevents indirect supervision from stronger policies (e.g., those obtained
via training-from-scratch) that the system aims to approximate.

\paragraph{Predictor model.}
We use a histogram-based gradient boosting regressor operating on policy
embeddings.
This model provides sufficient flexibility to capture differences in policy
behavior while remaining computationally efficient.

\begin{figure*}[h]
    \centering
    \begin{subfigure}[b]{0.32\textwidth}
        \includegraphics[width=\textwidth]{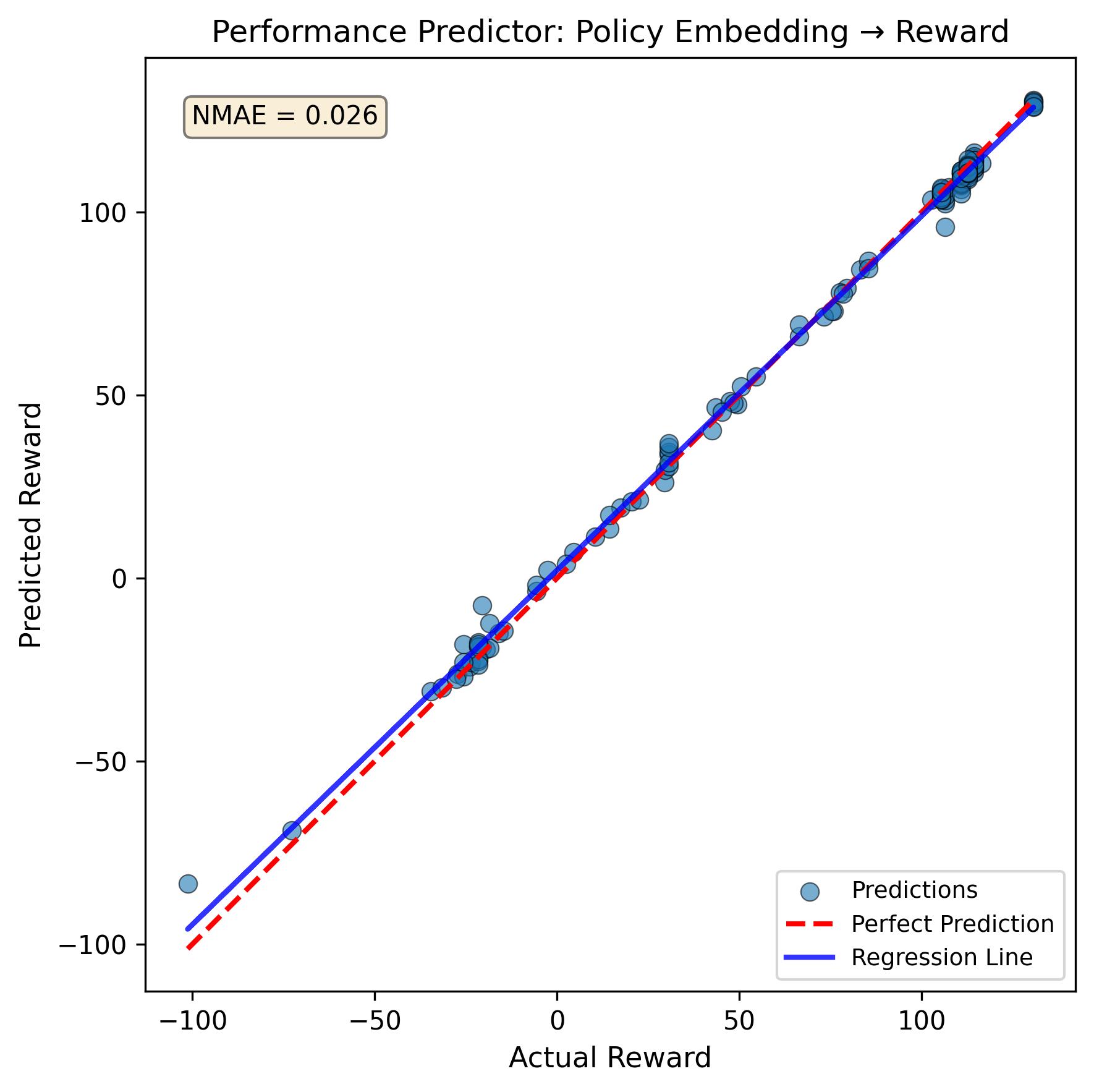}
        \caption{Trivial Predictor for X1}
        \label{fig:reg-base}
    \end{subfigure}
    \hfill
    \begin{subfigure}[b]{0.32\textwidth}
        \includegraphics[width=\textwidth]{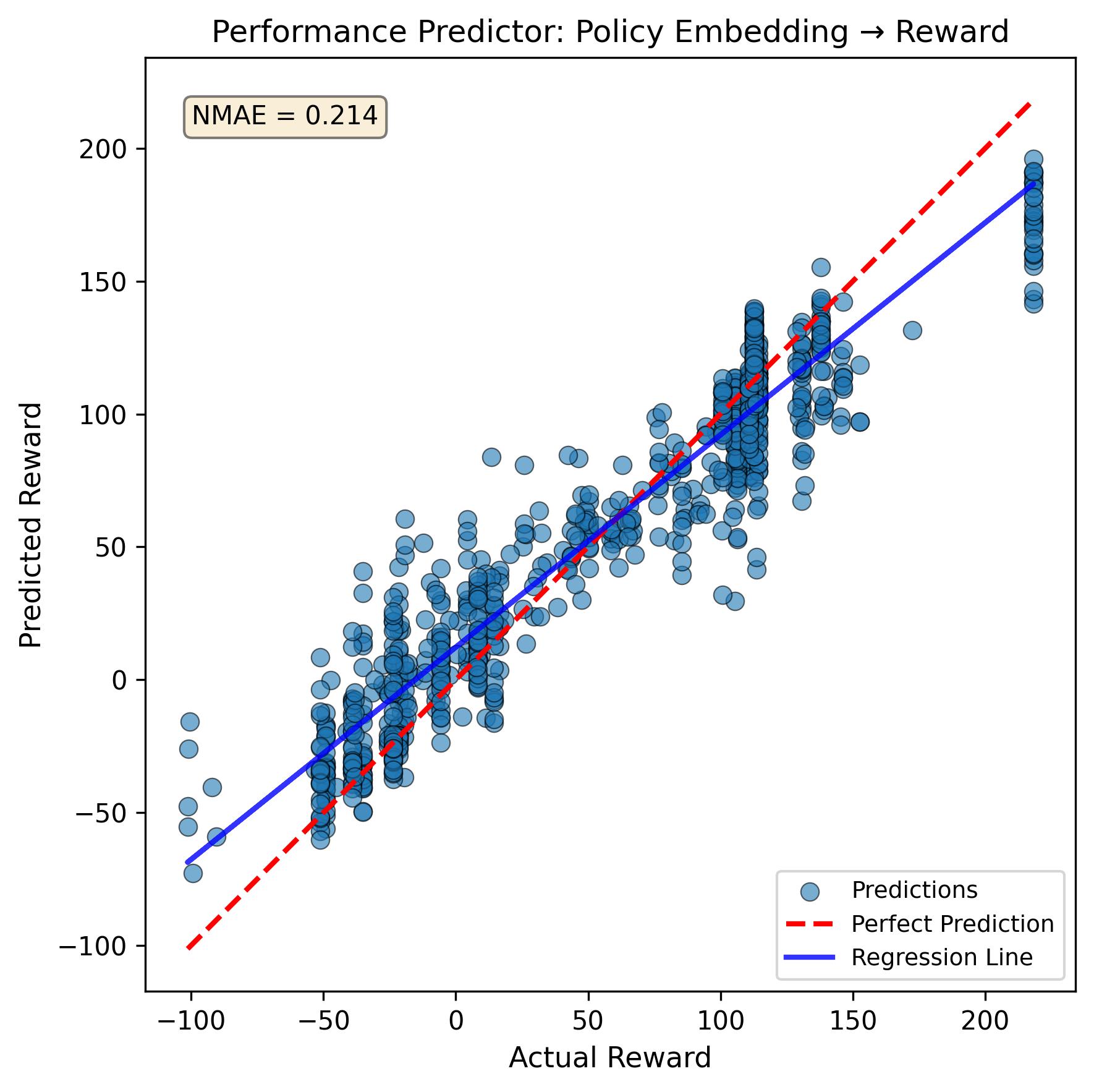}
        \caption{Double Predictor for X5}
        \label{fig:reg-pair}
    \end{subfigure}
    \hfill
    \begin{subfigure}[b]{0.32\textwidth}
        \includegraphics[width=\textwidth]{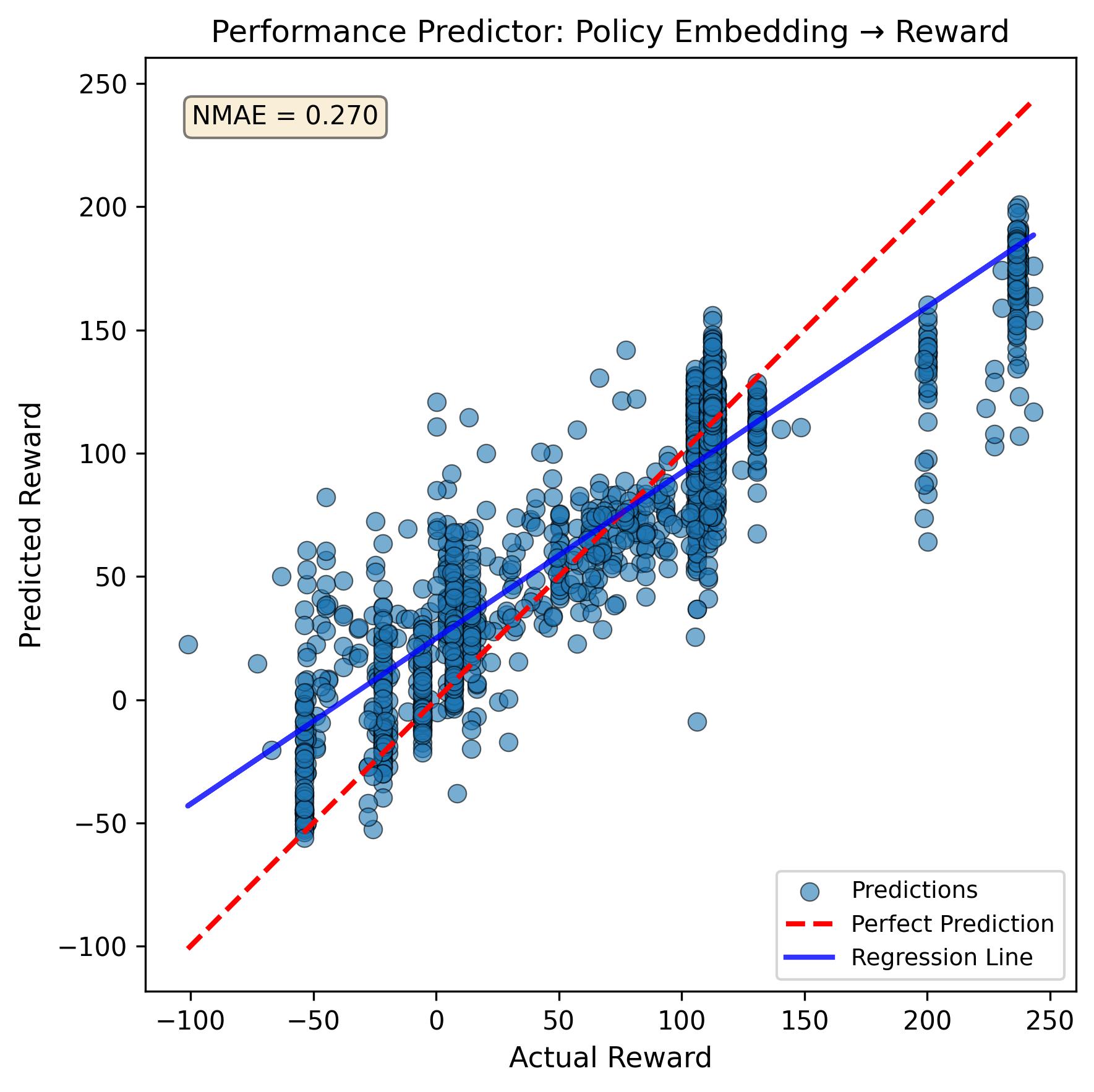}
        \caption{Triple Predictor for X10}
        \label{fig:reg-trip}
    \end{subfigure}
    \caption{Performance predictor fit for $16\times16$ ($\gamma=0$) across
different training horizons and budgets. The horizontal axis represents the
ground-truth reward, while the vertical axis shows the predicted reward using
the histogram-based gradient boosting regressor.}
    \label{fig:reg-triple-plot}
\end{figure*}

\subsection{Main Results: Return and Runtime}
\label{subsec:main-results}

We evaluate inference-time policy reuse along two complementary dimensions:
(i) \emph{policy quality}, measured by average episodic return under the
composite objective, and
(ii) \emph{computation cost}, measured as the offline time required to produce
the final policy.

We compare the three composition strategies (TC, HC, EC) to
Training-from-Scratch (TFS) across multiple training budgets ($X1$, $X5$, $X10$)
and grid sizes.

\paragraph{Support-limited regime ($\gamma = 0$).}
In the primary setting of this paper, inference-time reuse performs strongly.
Both TC and HC achieve returns that are competitive with, and in some cases
comparable to, TFS, while requiring substantially less computation.
This demonstrates that, when relevant transitions are present in the policy
library, offline composition can effectively replace retraining.

At the same time, the choice of composition strategy impacts performance.
TC provides the lowest computation cost but is sensitive to ranking errors,
while HC improves robustness by exploring a larger portion of the available
support.
EC can further improve performance in some cases, but at a significantly higher
computational cost.

\paragraph{Runtime comparison.}
Across all settings, inference-time reuse yields substantial reductions in
computation time compared to TFS.
This gap becomes more pronounced as the problem size or training budget
increases, highlighting the scalability advantage of offline reuse.

Overall, these results show that in the support-limited regime, inference-time
policy reuse can achieve a favorable trade-off between performance and
computation, provided that the policy library offers sufficient coverage.

\paragraph{Hybrid top-$k$ selection.}
HC introduces a top-$k$ parameter that controls how many candidate policies are
retained per subtask before composition.
When $k=1$, HC reduces to TC, selecting a single policy per subtask.
When $k$ includes all candidates, HC becomes equivalent to EC.
Thus, $k$ directly controls the extent to which the available support is
explored.

As $k$ increases from $1$, average return improves rapidly, reflecting the
benefit of considering multiple candidates and recovering combinations that are
missed by more restrictive strategies.
However, beyond a small value of $k$, performance gains quickly saturate while
offline computation continues to grow.

In both $8\times8$ and $16\times16$ environments, we observe that modest values
of $k$ already capture most of the achievable performance.
In some cases, composed policies match or exceed TFS, suggesting that reuse can
effectively leverage existing behaviors without requiring full retraining.

Based on this trade-off, we select $k=3$ as a practical operating point, which
provides a good balance between performance and computation while avoiding the
cost of more exhaustive exploration.

\begin{figure}[ht]
    \centering
    \includegraphics[width=0.95\linewidth]{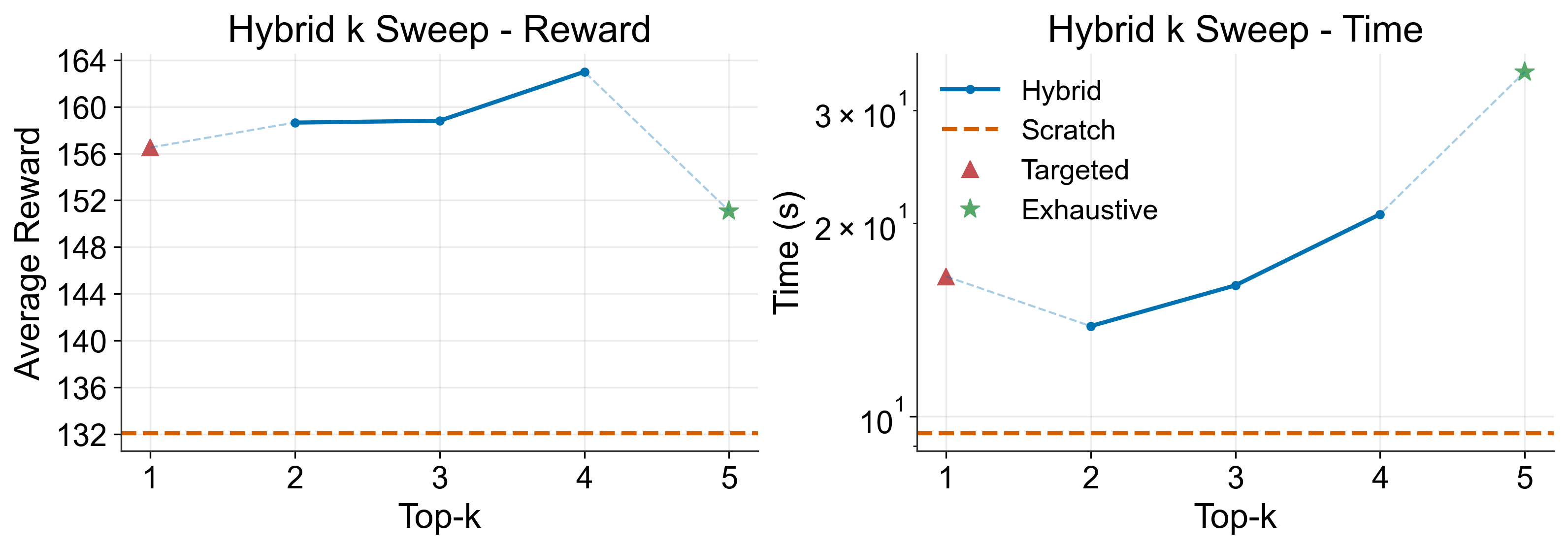}\\
    \vspace{0.4em}
    \includegraphics[width=0.95\linewidth]{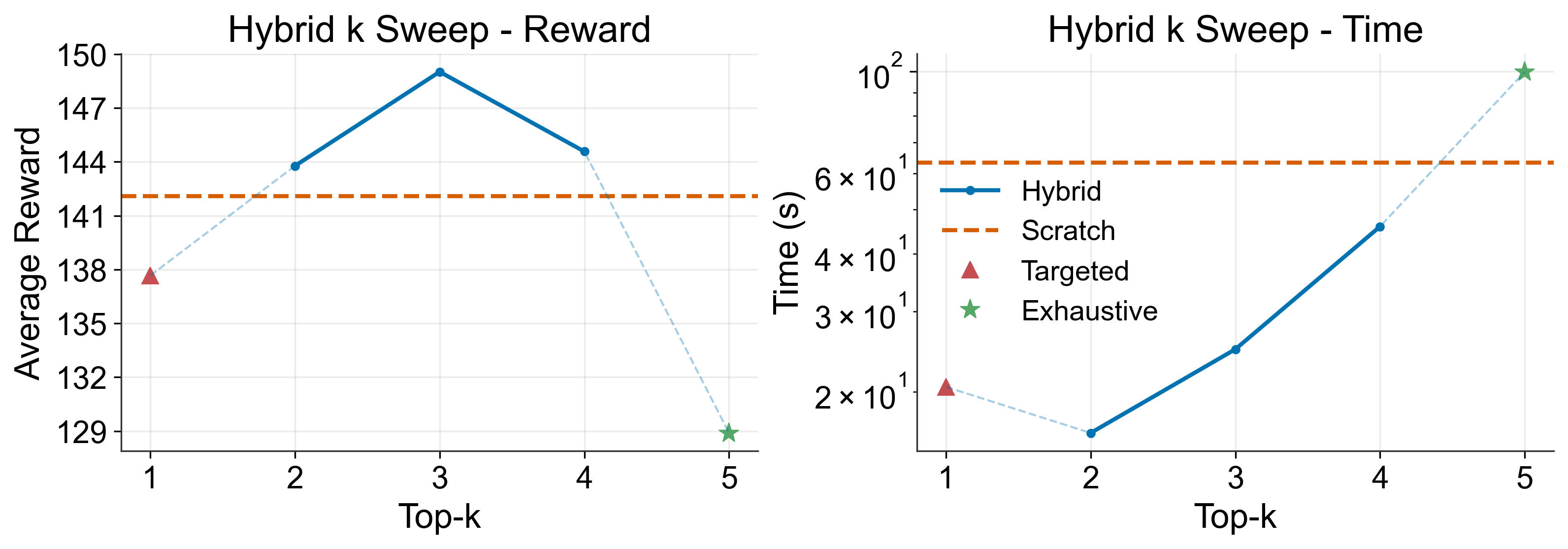}
    \caption{Hybrid top-$k$ sweep. Top: $8\times8$; bottom: $16\times16$.}
    \label{fig:hybrid-sweep}
\end{figure}

\begin{figure}[ht]
    \centering
    \textbf{Tabular, $\gamma=0$}\\[0.3em]
    \includegraphics[width=0.95\linewidth]{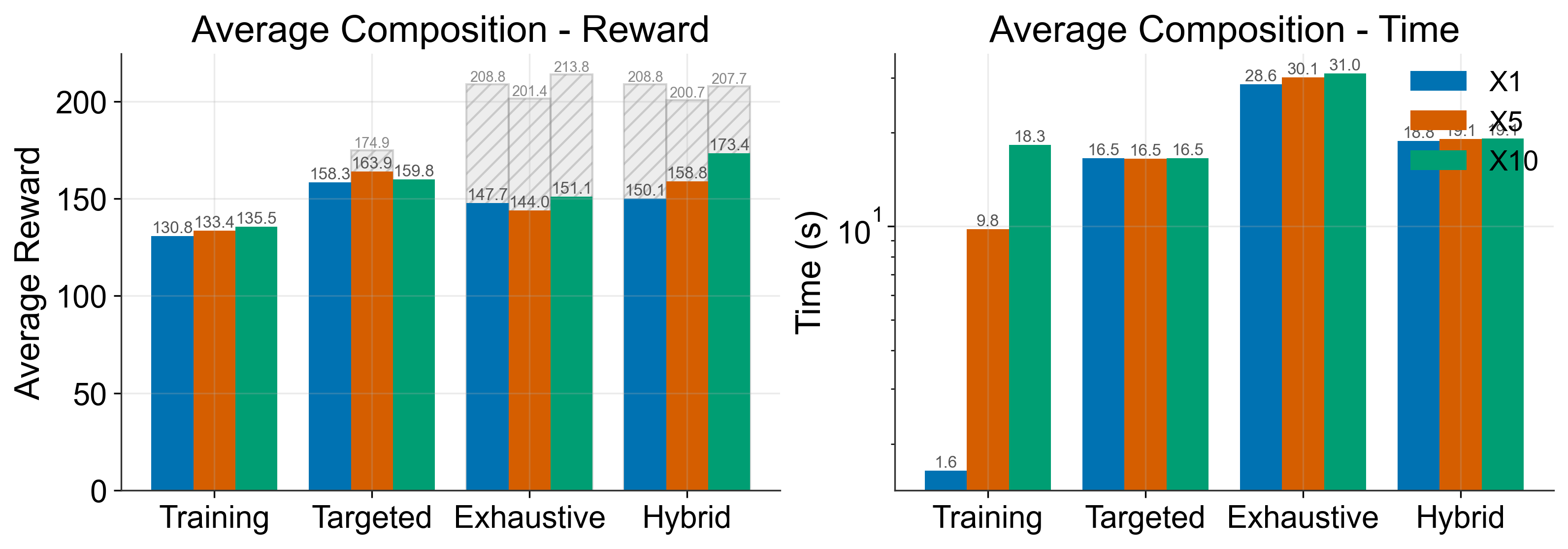}\\
    \vspace{0.4em}
    \includegraphics[width=0.95\linewidth]{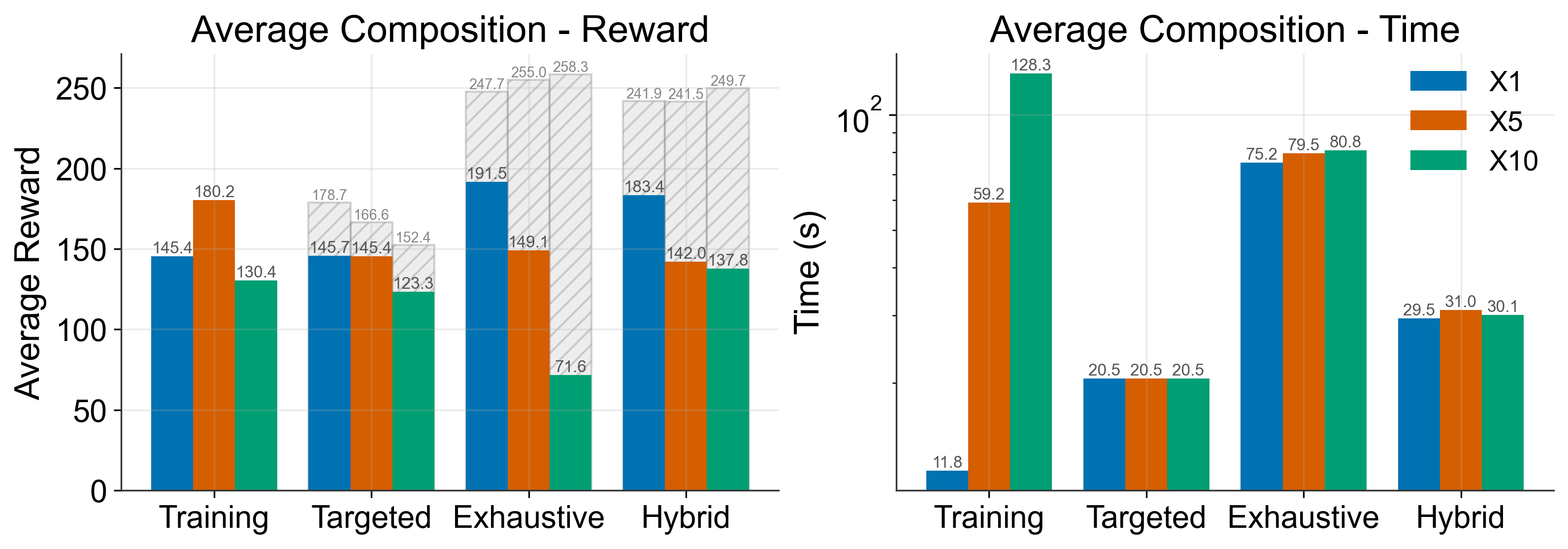}
    \caption{
    Average episodic return (left) and offline computation time (right) for
    TFS, TC, EC, and HC across $X1$, $X5$, and $X10$ in the tabular
    $\gamma=0$ setting.
    Top: $8\times8$; bottom: $16\times16$.
    TFS runtime grows substantially with training budget, while TC/EC/HC
    remain nearly constant and achieve comparable return.
    }
    \label{fig:return-time-tabular-0}
\end{figure}

\begin{figure}[ht]
    \centering
    \textbf{Tabular, $\gamma=0.99$}\\[0.3em]
    \includegraphics[width=0.95\linewidth]{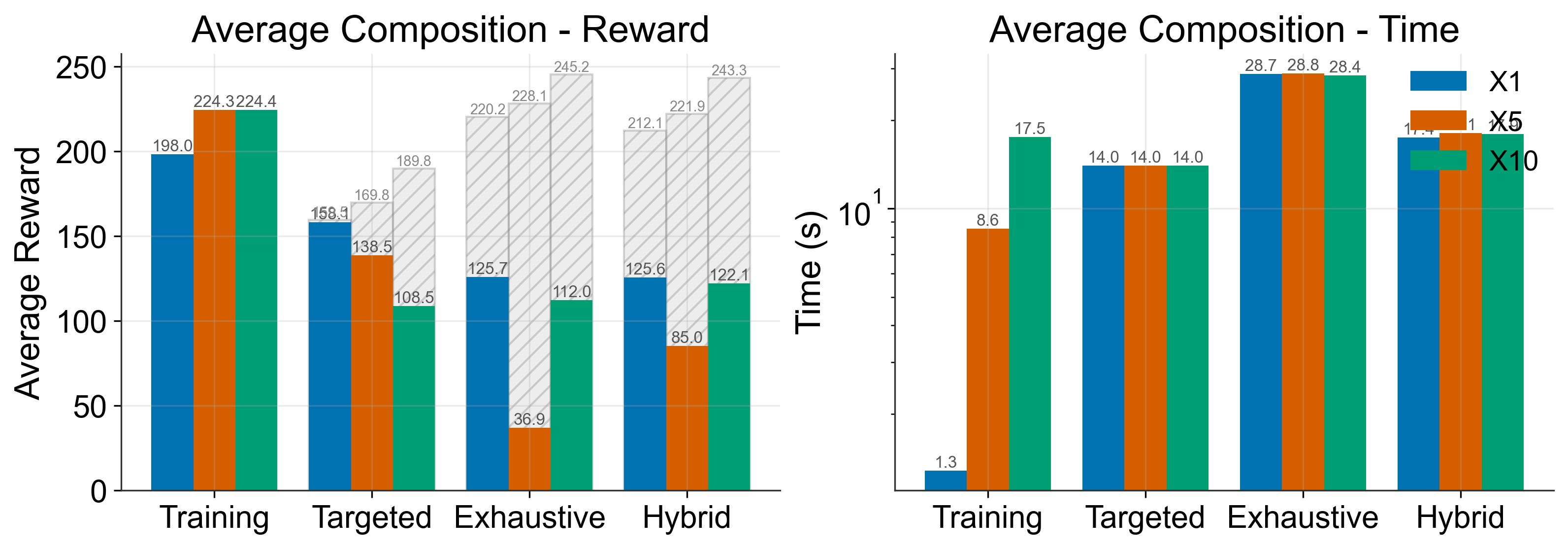}\\
    \vspace{0.4em}
    \includegraphics[width=0.95\linewidth]{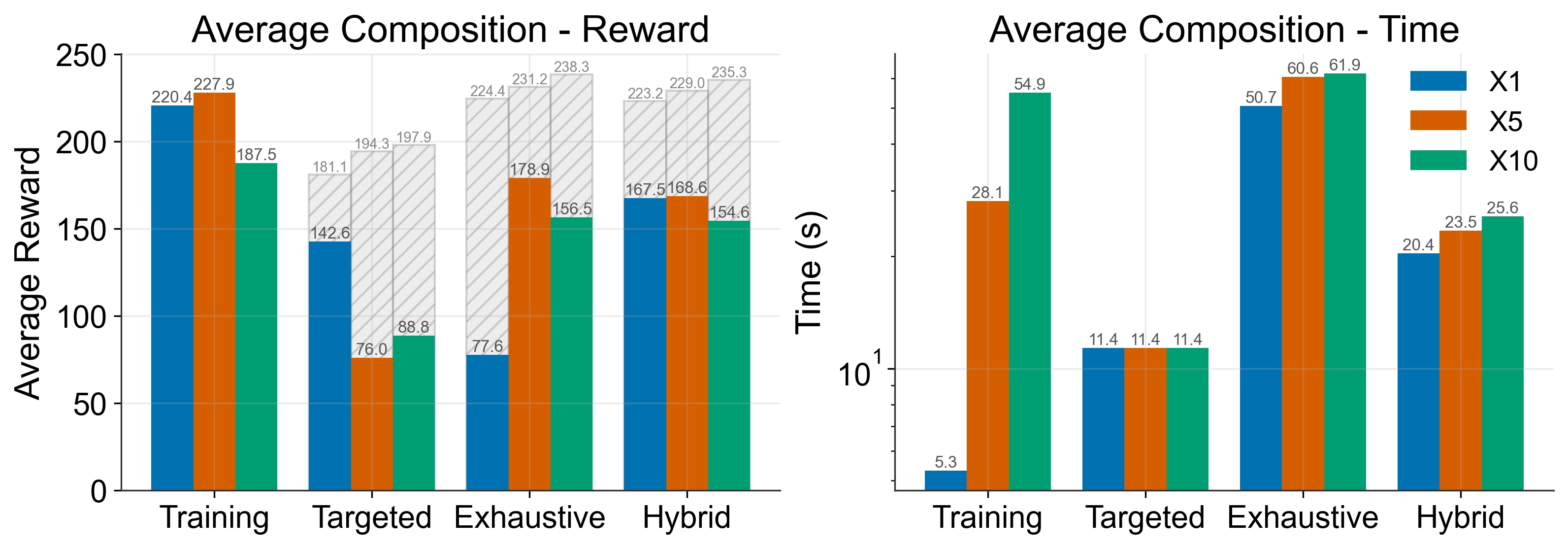}
    \caption{
    Average episodic return (left) and offline computation time (right) for
    TFS, TC, EC, and HC across $X1$, $X5$, and $X10$ in the tabular
    $\gamma=0.99$ setting, where the support-limited composition operator is
    applied to policies trained outside its theoretical assumptions.
    Top: $8\times8$; bottom: $16\times16$.
    }
    \label{fig:return-time-tabular-99}
\end{figure}

\begin{figure}[ht]
    \centering
    \textbf{DQN}\\[0.3em]
    \includegraphics[width=0.95\linewidth]{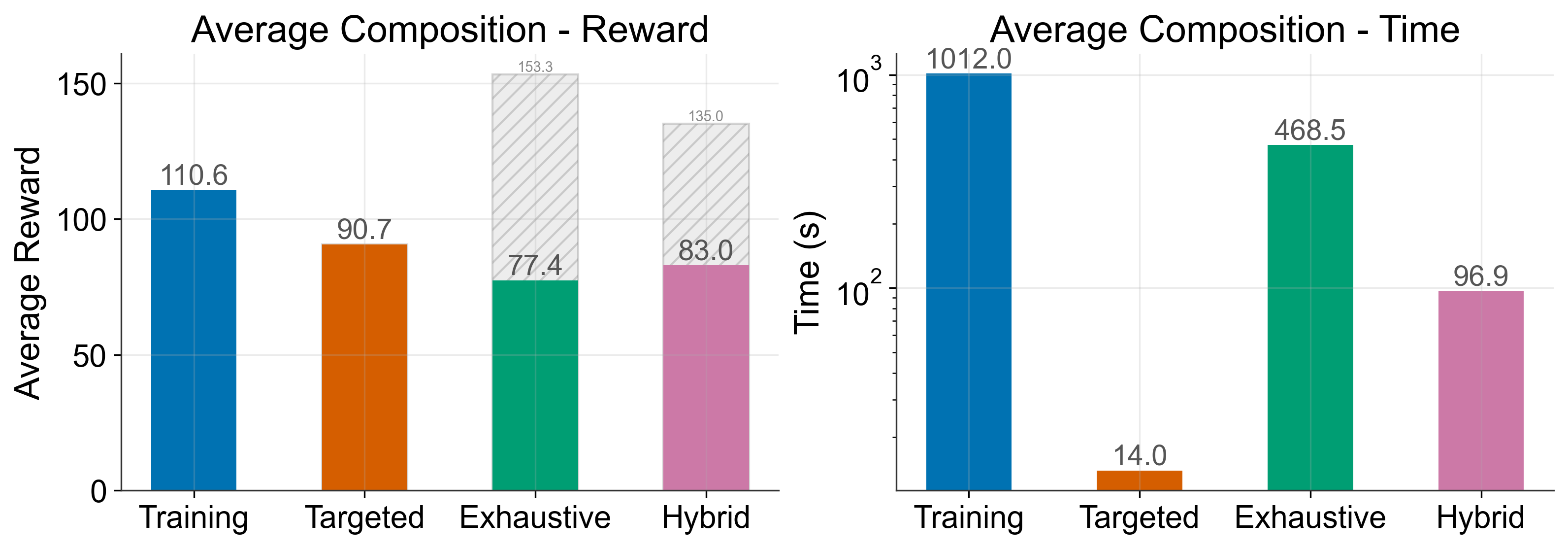}\\
    \vspace{0.4em}
    \includegraphics[width=0.95\linewidth]{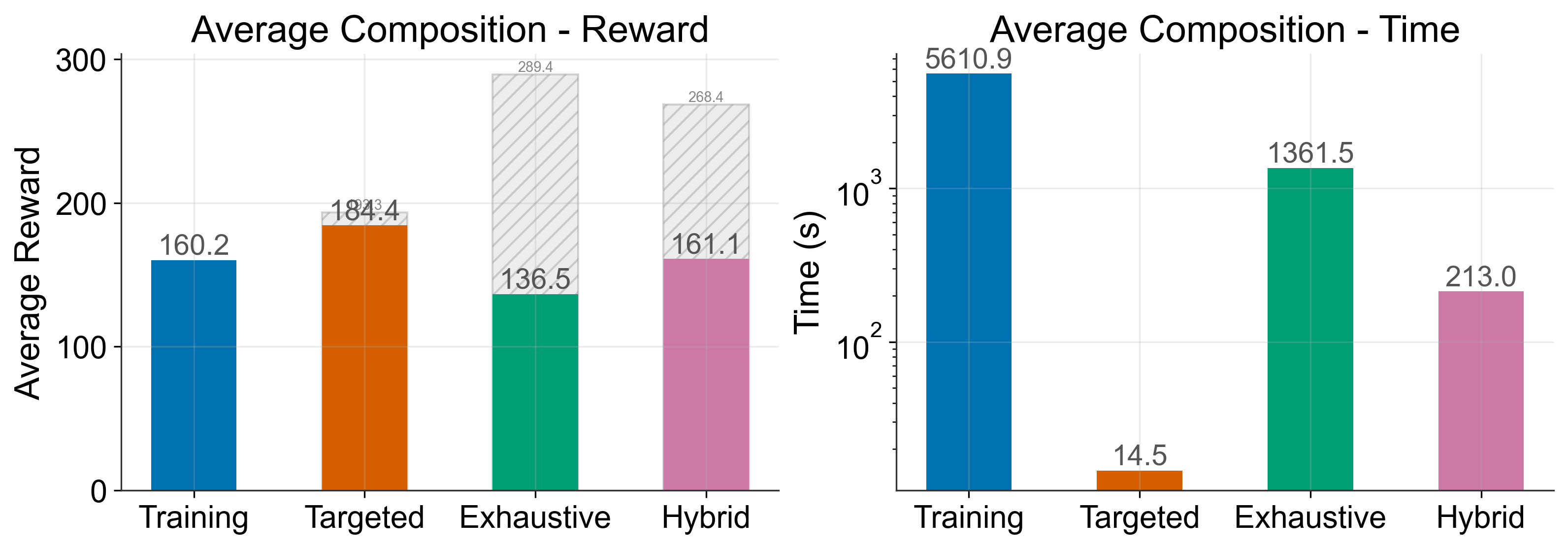}
    \caption{
    Average episodic return (left) and offline composition time (right) for
    TFS, TC, EC, and HC using DQN policies composed via GPI
    (Eq.~\ref{eq:gpi_dqn}).
    Top: $8\times8$; bottom: $16\times16$.
    }
    \label{fig:return-time-dqn}
\end{figure}

\begin{figure}[ht]
    \centering
    \textbf{PPO}\\[0.3em]
    \includegraphics[width=0.95\linewidth]{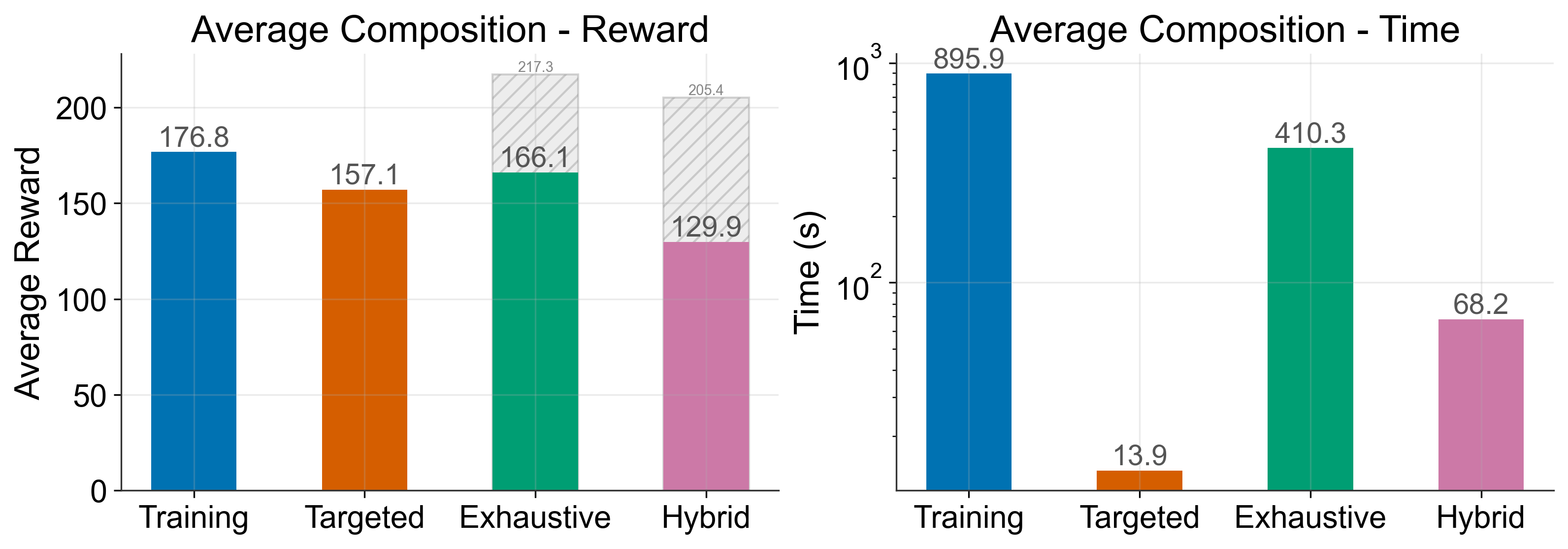}
    \caption{
    Average episodic return (left) and offline composition time (right) for
    TFS, TC, EC, and HC using PPO policies composed via critic-based GPI
    (Eq.~\ref{eq:gpi_ppo}) on the $8\times8$ grid.
    }
    \label{fig:return-time-ppo}
\end{figure}

Figures~\ref{fig:return-time-tabular-0} and~\ref{fig:return-time-tabular-99}
summarize the main comparison between composition strategies in the tabular
setting.
Figure~\ref{fig:return-time-tabular-0} reports results for the
support-limited regime ($\gamma=0$), while
Figure~\ref{fig:return-time-tabular-99} shows results when policies are trained
with $\gamma=0.99$, i.e., outside the assumptions of the composition operator.
Figures~\ref{fig:return-time-dqn} and~\ref{fig:return-time-ppo} extend the
comparison to deep RL settings (DQN and PPO).
In all figures, the left column reports average episodic return and the right
column reports end-to-end wall-clock time.

\paragraph{Return comparison.}
In the support-limited regime ($\gamma=0$), inference-time reuse performs
strongly.
Both TC and HC achieve returns close to TFS across training budgets, and in some
cases match or exceed it.
This confirms that when relevant transitions are present in the policy library,
offline composition can effectively recover high-quality policies.

Each figure also reports an \textbf{upper bound}, obtained by directly evaluating
all available policy snapshots on the composite task.
This upper bound represents the best performance achievable given the available
support.
The gap between this bound and the pipeline results reflects imperfections in
retrieval and ranking, rather than a limitation of the composition operator
itself.

Outside the support-limited regime ($\gamma>0$), performance degrades.
In the tabular $\gamma=0.99$ setting, composition becomes more sensitive to
approximation errors, as long-horizon effects cannot be propagated by the
support-limited operator.
Similarly, in deep RL settings, composed policies remain competitive but
consistently fall below the upper bound, reflecting the combined effect of
approximate composition (via GPI) and imperfect ranking.

For DQN, TC achieves performance close to TFS and approaches the upper bound in
larger environments, while HC and EC provide marginal improvements at increased
cost.
PPO results follow a similar qualitative pattern.

Overall, these results highlight a clear distinction:
inference-time reuse is effective when the problem is support-limited, but
becomes less reliable when value propagation or generalization beyond observed
transitions is required.

\paragraph{Runtime comparison.}
Training-from-Scratch (TFS) becomes increasingly expensive as environment size
and training budget grow, whereas inference-time reuse operates on a fixed
policy library and remains largely insensitive to training horizon.
Among reuse strategies, TC is consistently the fastest, HC provides a moderate
increase in cost, and EC is the most expensive due to combinatorial
enumeration.

In practice, EC often incurs higher cost than TFS in multi-objective settings,
making it impractical despite occasional performance gains.
In contrast, TC and HC provide substantial computational savings.
This advantage is particularly pronounced in deep RL, where training costs are
high: for example, in DQN on $16\times16$, TC and HC reduce wall-clock time by
orders of magnitude compared to TFS.

\paragraph{Hybrid efficiency ($k=3$).}
Hybrid Composition (HC) provides a favorable trade-off between performance and
computation.
As problem size increases, its relative efficiency improves: while HC may be
slower than TFS on small grids, it becomes significantly faster on larger ones.
This reflects the different scaling behaviors of the two approaches:
training cost grows rapidly with problem size, whereas composition cost grows
much more slowly.

In deep RL settings, this advantage appears even earlier, with HC already
outperforming TFS in runtime on smaller environments.

\paragraph{Core findings.}
Across all settings, we observe three main patterns.

\textbf{(1) Reuse is effective in the support-limited regime.}
When $\gamma = 0$, inference-time composition achieves performance close to
TFS, confirming that reuse can recover high-quality policies when the necessary
transitions are present in the policy library.

\textbf{(2) Reuse degrades outside this regime.}
When $\gamma > 0$, performance becomes more sensitive to approximation and
coverage.
Since value propagation cannot be performed, composition relies on incomplete
information, leading to a consistent gap with the upper bound.

\textbf{(3) Exploration of the policy space matters.}
Increasing the breadth of exploration (from TC to HC to EC) improves the chance
of recovering useful combinations, but at a rapidly increasing computational
cost.
HC provides the most practical balance, capturing most of the achievable
performance while remaining efficient.

Overall, these results show that the effectiveness of inference-time policy
reuse is primarily determined by the availability of sufficient support and the
extent to which it is explored, rather than by the specific composition
mechanism.

\subsection{Experimental Details}
\label{subsec:exp-details}

\paragraph{Environment.}
We evaluate on deterministic GridWorld environments of size $8\times8$ and
$16\times16$.
Each layout contains a start cell, an exit cell, gold cells, obstacle cells,
hazard cells, and a lever cell.
For each random seed, we sample one layout and reuse it across all reward
settings to ensure comparability.
Layouts are accepted only if at least one hazard-free path exists from start to
exit and from start to lever to exit.

\paragraph{Reward components.}
We define four base objectives: \texttt{path}, \texttt{gold},
\texttt{hazard}, and \texttt{lever}, each with dense shaping.
The \texttt{path} reward encourages progress toward the exit and gives a
terminal bonus on success.
The \texttt{gold} reward encourages collection of gold items and rewards exit
only when collection is complete.
The \texttt{hazard} reward penalizes contact with hazards and encourages safer
navigation.
The \texttt{lever} reward encourages reaching the lever before exiting.
Composite tasks are formed by summing active reward components, e.g.,
\texttt{path-gold}, \texttt{path-gold-hazard}, and
\texttt{path-gold-hazard-lever}.
Each active objective also contributes a small step cost, so multi-objective
tasks are more time-sensitive.

\paragraph{Action space and episode termination.}
The agent has four deterministic actions (up, down, left, right).
Invalid moves leave the agent in place.
Episodes terminate when the agent reaches the exit, steps on a hazard, or
reaches a horizon of $N^2$ steps.

\paragraph{Training-from-Scratch baseline.}
In the tabular setting, the Training-from-Scratch (TFS) baseline uses SARSA
with learning rate $\alpha=0.1$ and an $\varepsilon$-greedy exploration
schedule decaying from $1.0$ to $0.01$.
We consider three training budgets:
$X1$ (10k episodes), $X5$ (50k episodes), and $X10$ (100k episodes),
with decay rates adjusted to maintain comparable exploration profiles.
The discount factor follows the experimental regime:
$\gamma=0$ in the support-limited setting and $\gamma=0.99$ in the
out-of-regime setting.

\paragraph{Policy library and diversity.}
To support offline evaluation, we build a diverse policy library by varying
random layouts, training budget, and checkpoint quality.
For each training run, we retain three reward-stratified checkpoints
(\emph{best}, \emph{mid}, \emph{low}), yielding policies with varying behavior
quality.
This diversity is used both for retrieval/composition and for training the
offline predictor.

\paragraph{Policy evaluation and predictor.}
Candidate policies are represented using $\pi^2$VEC-style behavioral embeddings
computed from a structured state representation over GridWorld states.
To rank policies without environment interaction, we train an offline predictor
$f(\Psi_\pi)$ mapping embeddings to estimated return.
We use a histogram-based gradient boosting regressor, which offered the best
practical trade-off between expressiveness and efficiency among the lightweight
models we tested.
To avoid temporal leakage, predictors are trained only on policies that would be
available at the corresponding composition stage and training budget.

\paragraph{Deep RL settings.}
To test generality beyond tabular policies, we also evaluate DQN and PPO on the
\texttt{LeverGrid} MiniGrid environment.
Both use a convolutional architecture with a local $7\times7$ observation.
For DQN, composition is performed via Generalised Policy Improvement (GPI) over
per-action Q-values.
For PPO, composition delegates action selection to the policy whose critic
assigns the highest state value.
These deep compositions are fully offline but should be viewed as approximate
rather than exact Q-value composition.

\subsection{Discussion}
\label{sec:disc}

\noindent\textbf{Performance--scalability trade-offs.}
Our results show that composition strategies differ primarily in how they
explore the available support and how their cost scales, rather than in their
peak achievable performance.
Targeted Composition (TC) is efficient but sensitive to selection errors,
while Exhaustive Composition (EC) can recover stronger solutions at a
prohibitive computational cost.
Hybrid Composition (HC) provides the most practical balance, capturing most of
the achievable performance while remaining scalable.
These results highlight that the key challenge is not the composition operator
itself, but how effectively the available support is explored under a limited
computational budget.

\noindent\textbf{Role of regimes and guarantees.}
The effectiveness of inference-time reuse depends critically on the underlying
regime.
In the support-limited setting ($\gamma = 0$), composition performs reliably and
closely matches the best achievable performance given the available policies.
When applied outside this regime ($\gamma > 0$), performance degrades, as
long-horizon dependencies cannot be recovered from the available support.
This gap reflects a fundamental limitation of offline reuse rather than a
failure of a specific composition strategy.

\noindent\textbf{Sources of error.}
Two main factors limit performance in practice.
First, policy selection depends on approximate ranking based on offline
information, which can lead to suboptimal choices.
Second, the policy library may not provide sufficient coverage of the state and
action space required by the composite task.
These limitations become more pronounced as tasks require coordination across
multiple objectives or longer horizons.

\noindent\textbf{Generality of the framework.}
The \textsc{lever} framework is largely independent of the specific composition
mechanism.
Our experiments demonstrate that it can accommodate different composition
operators, including tabular Q-value summation and approximate methods such as
GPI for deep RL.
This suggests that the framework can be extended to other policy representations,
provided that candidate policies can be evaluated and compared offline.

\noindent\textbf{Limitations and future directions.}
A key limitation of our approach is its reliance on the availability and
diversity of the policy library.
When relevant behaviors are not present, composition cannot recover them.
In addition, offline evaluation remains approximate, and improving the
generalisation of policy embeddings and predictors is an important direction
for future work.
Finally, some composition strategies require access to environment dynamics for
offline evaluation, which may not be available in general settings.

\section{Conclusion}

We studied inference-time policy reuse, where a fixed library of pre-trained
policies must be leveraged to solve new tasks without additional environment
interaction.
Using the \textsc{lever} framework as a practical instantiation, we showed that
the effectiveness of reuse depends fundamentally on the underlying regime.

In the support-limited setting ($\gamma = 0$), inference-time composition can
recover policies with performance close to training from scratch while
providing substantial computational savings.
In contrast, when long-horizon dependencies are present ($\gamma > 0$),
performance degrades, as value propagation cannot be recovered from the
available support.
This highlights that the main limitation of offline reuse is not the
composition mechanism itself, but the availability and coverage of relevant
transitions.

Our results further show that exploring the policy space improves performance,
but introduces a trade-off between coverage and computational cost.
Hybrid Composition provides the most practical balance, capturing most of the
achievable performance while remaining efficient.

More broadly, this work suggests that inference-time policy reuse is a viable
alternative to retraining in settings where sufficient support is available,
but becomes unreliable when generalisation beyond observed transitions is
required.
Improving offline evaluation and representation learning for policies remains a
key direction for extending reuse to more complex settings.

\section*{Impact Statement}

This work contributes to the study of model reusability in reinforcement
learning, with a focus on inference-time policy reuse.
By enabling the composition of pre-trained policies without additional
training or environment interaction, our approach has the potential to reduce
the computational cost and expertise required to deploy RL systems.
This could make decision-making systems more accessible and energy-efficient,
particularly in settings where retraining is expensive or impractical.

At the same time, the ability to reuse and compose policies raises important
considerations.
Since decisions are made based on previously learned behaviors, the quality and
coverage of the policy library become critical.
Incomplete or biased policy collections may lead to suboptimal or unsafe
decisions, especially when reused in new contexts.
In addition, the absence of online interaction limits the system's ability to
adapt or correct errors.

Overall, our work highlights both the opportunities and limitations of
inference-time reuse.
We believe that improving the robustness of offline evaluation and ensuring
diverse, well-curated policy libraries are key steps toward the responsible
deployment of such systems.

\end{document}